\documentclass[letterpaper]{article} 
\usepackage{aaai25}  
\usepackage{times}  
\usepackage{helvet}  
\usepackage{courier}  
\usepackage[hyphens]{url}  
\usepackage{graphicx} 
\usepackage{natbib}  
\usepackage{caption} 
\usepackage{algorithm}
\usepackage{graphicx}
\usepackage{algpseudocode}
\usepackage{tablefootnote}
\usepackage{footmisc}
\usepackage{makecell}
\usepackage{multirow}
\usepackage{amsmath}

\usepackage{newfloat}
\usepackage{listings}
\DeclareCaptionStyle{ruled}{labelfont=normalfont,labelsep=colon,strut=off} 
\floatstyle{ruled}
\newfloat{listing}{tb}{lst}{}
\floatname{listing}{Listing}
\title{Generative Video Diffusion for Unseen Novel Semantic Video Moment Retrieval}
\author{
Dezhao Luo\textsuperscript{\rm 1},
Shaogang Gong\textsuperscript{\rm 1}\thanks{Corresponding authors}  , 
Jiabo Huang\textsuperscript{\rm 2}, 
Hailin Jin\textsuperscript{\rm 3}, 
and Yang Liu\textsuperscript{\rm 4,5}\footnotemark[1]
}
\affiliations {
    \textsuperscript{\rm 1}Queen Mary University of London  \\\textsuperscript{\rm 2}Sony AI \\ 
    \textsuperscript{\rm 3}Adobe Research\\
    \textsuperscript{\rm 4}WICT, Peking University  \\
    \textsuperscript{\rm 5}State Key Laboratory of General Artificial Intelligence, Peking University
    \\
    \{dezhao.luo, s.gong\}@qmul.ac.uk, raymond.huang@sony.com, hljin@adobe.com, yangliu@pku.edu.cn
}

\usepackage{xspace}
\def\abbrname{FVE\xspace}
\newcommand{\eg}{e.g.,\xspace}
\usepackage{amsmath} 
\usepackage{amssymb} 
\usepackage{amsfonts}
\usepackage{xcolor}
\newcommand{\dz}[1]{{\color{black}#1}}
\newcommand{\sg}[1]{{\color{black}#1}}

\usepackage{float}

\begin{document}

\maketitle
\begin{abstract}
Video moment retrieval (VMR) aims to locate the most likely
  video moment(s) corresponding to a text query in
untrimmed videos. Training of existing methods is limited by
  the lack of diverse and generalisable VMR datasets, hindering
their ability to generalise moment-text associations to
queries containing novel semantic concepts (unseen both visually
  and textually in a training source domain). For model generalisation to novel semantics, existing methods rely heavily on assuming to have access to both video and text sentence pairs from a
  target domain in addition to the source domain pair-wise training data. This is neither practical nor scalable.
In this work, we introduce a more generalisable approach by
  assuming {\em only} text sentences describing new semantics are available in model training {\em without} having seen any videos from a target domain. To that end, we propose a Fine-grained Video Editing framework, termed 
  \abbrname, that explores generative video
diffusion to facilitate fine-grained video editing from the seen
  source concepts to the unseen target sentences consisting of new
  concepts. This enables generative hypotheses of unseen video
  moments corresponding to the novel concepts in the target
  domain. This fine-grained generative video diffusion  
retains the original video structure and subject specifics from the
source domain while introducing semantic distinctions of unseen
novel vocabularies in the target domain. A critical challenge
  is how to enable this generative fine-grained diffusion process to
  be meaningful in optimising VMR, more than just synthesising
  visually pleasing videos. We solve this problem by introducing a
  hybrid selection mechanism that integrates three quantitative
  metrics to selectively incorporate synthetic video 
moments (novel video hypotheses) as enlarged additions to the
  original source training data, whilst minimising potential
detrimental noise or unnecessary repetitions in the novel synthetic
videos harmful to VMR learning. Experiments on three datasets
demonstrate the effectiveness of \abbrname to unseen novel semantic
video moment retrieval tasks.
\end{abstract}

\section{Introduction}

Given an untrimmed video and a sentence query, video moment
  retrieval (VMR) aims to locate the most relevant video moment(s)
    semantically corresponding to the query. This task is
challenging because it requires extracting semantic
associations between visual and textual data with precise time
  locations. Annotating VMR datasets requires indexing temporal video
moments with corresponding 
sentences and distinguishing
them from contextual moments within the video, which is a more
intricate and less scalable process compared to labelling image-text
or video-text pairs. 

Due to the lack of large-scale video moment-text datasets, VMR
models are struggling to learn generalisable {\em novel} moment-text
associations beyond the training source domains,
resulting in an inferior cross-domain adaptation where training and testing data display biases. 
\dz{In contrast to the biases in the moment
location or length \cite{canshuffle},} we tackle a more intricate challenge: semantic
biases across domains, where the
semantic in the testing domain is {\em novel} to the training domain. By `novel', it means novel text vocabularies and
  their corresponding video moments both unseen in source domain
  training. \dz{ Comparing to previous methods employing co-training strategies on both
source and target datasets \cite{eva}, and other methods~\cite{psvl}
creating pseudo moment-text associations by generating textual descriptions for a target video, we propose a more scalable and accessible approach to 
learning generalisable VMR to unseen novel semantics by exploring
target domain sentences describing new semantics only,
  without any videos from the target domain.}

Recent successes in generative diffusion models \cite{stable-diffusion,wang2023modelscope} have
demonstrated the power of hypothesising new holistic videos with text
prompts. To benefit novel semantic VMR, a potential approach \dz{is to generate videos conditioned on both source video moments and a target
sentence of novel semantics/concepts.}
 However, there are several non-trivial challenges in
synthesising a meaningful visual hypothesis for a VMR video
displaying the same subject performing different actions in a
  similar environment (background). The first challenge lies in regulating the generation of a video moment based on novel semantics referenced in a target domain sentence while retaining other contextual information intact. \dz{ Existing text conditioned video generation \cite{wang2023modelscope} or editing methods \cite{TAV}  lack specific constraints  or image conditioned methods~\cite{jiang2024videobooth}, \sg{leading towards} inaccurate
subject details~(shown in Table \ref{table:quan}, Fig. \ref{fig:qualitative} and the Supplementary)}. This highlights the need
for an {\em instance-preserving video action editing} method
  constrained by accurate subject specifics~(fine-grained
  details).   The second challenge is that existing generative
    video editing techniques \cite{TAV,videop2p}
    heavily depend on manual intervention to choose appropriate
    editing text prompts to `guide' credible outcomes.
    Automated video generation poses a risk of producing implausible
    or trivial repetitive videos, not meaningful and
      potentially detrimental to the generalisability of a VMR model
      if trained with such data. \sg{An unsolved critical problem of
        existing methods is how to select
     video hypothesis generation that can optimise VMR model learning to novel semantic concepts.}

To address these challenges, we introduce a Fine-grained Video Editing framework (\abbrname)
    which explores {\em fine-grained} generative video diffusion to
  finely edit videos of seen semantics from the source domain,
  guided by target sentences of new unseen semantics, thus
  hypothesising more meaningful unseen target video moments featuring these novel semantics for VMR training.
 To address the first challenge of moment generation by involving
 only `sentence-referred' local visual
variations while maintaining the background and the subject details
from the original source video,
we design a 2-stage video editing model
for simultaneous accurate subject preservation and
  fine-grained detail change guided by unseen novel concepts. Specifically, we first train an image diffusion model to align a text token with instances present across
a set of video frames, ensuring precise visual-textual alignment. Second, we treat video frames as a sequence and introduce a temporal layer within the image diffusion model to \dz{learn} the video motions. 
To tackle the challenge of minimising potential noisy
  and/or trivial repetitions in synthesising towards VMR
  training, we formulate three quantitative
  metrics aimed at filtering out implausible samples while selecting
  beneficial data for training the VMR model. Firstly, we introduce a
  cross-modal relevance metric to assess the semantic relevance
  between the target prompt and the generated video moment, thereby
  ensuring the quality of the moment-text association. Secondly, a
  uni-modal structure metric is introduced to evaluate the visual
  similarity of video moments between the source and generated
  moments before their utilisation in VMR training, which provides
  insights into video fidelity. Lastly, we introduce the model
  performance disparity metric, emphasising the importance of
  enriching VMR training by selecting more diverse synthesised hypotheses rather than
  duplicating similar visual content to the source domain videos. In practice,
  a synthetic video moment is incorporated into training only if it exhibits
  inferior retrieval performance, i.e. selection by a VMR
    discriminative constraint measured by a synthetic video's model performance
    disparity being high. 

We make three contributions:
(1) Instead of collecting moment-query pairs with novel semantic
  associations for VMR training in every new target domain in
    order to tackle the semantic bias across domains, we propose to
  {\em only} leverage target domain sentences containing unseen
    new semantics without any videos from the target domain. (2) We
  propose a Fine-grained Video Editing 
    framework~(\abbrname) to adapt a source domain video according to a target
    sentence of novel concepts as
  a controller for editing source domain videos to synthesise 
target moment-text associations for VMR training. Specifically, we
enhance generation control with an instance-preserving video diffusion
model, addressing the limitations of existing methods in maintaining
video subjects while altering action details. Additionally, we propose
a hybrid data selection strategy to curate the most beneficial
simulated videos for VMR training. (3)~We demonstrate the
effectiveness of \abbrname on both VMR and action editing tasks under diverse datasets.

\begin{figure*}[ht]
    \centering
    \includegraphics[width=\linewidth]{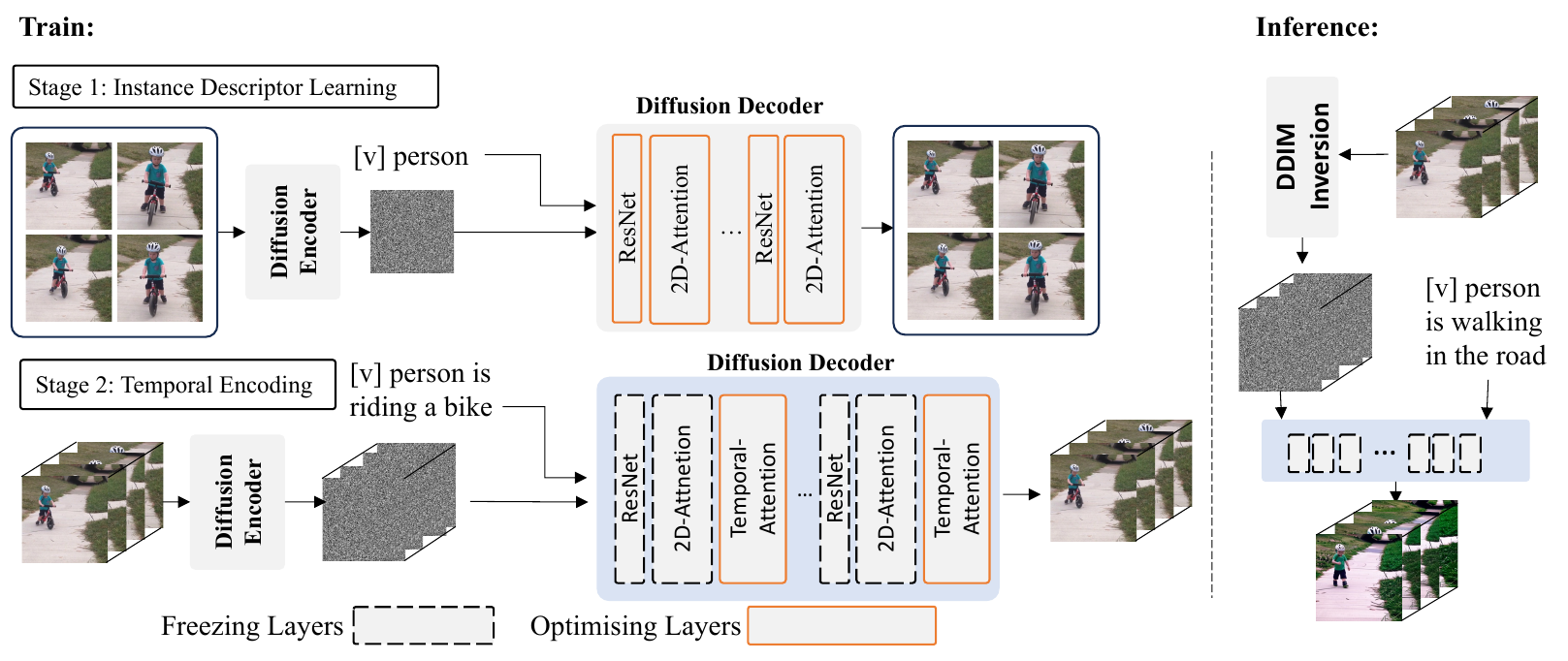}
    \caption{Our designed instance-preserving action editing model. We first take the video as a set of images and train an image diffusion model to align a special text token with the instance shared between those frames. Subsequently, we take those frames as a sequence and freeze the layers in the image diffusion model, and append a temporal layer to capture the video motions. }
    \label{fig:method}
\end{figure*}

\section{Related Works}
\noindent \textbf{Cross-Domain Video Moment Retrieval.}
To solve the problem of lacking an extensive video  dataset to train an effective generalisable VMR model,
CanShuffle \cite{canshuffle} proposed a data augmentation strategy to
solve the temporal bias problem by sacrificing the temporal semantics~\cite{cai2024semantic}, lacking the ability to
  understand novel semantics inherited in a video moment and its
  corresponding description. For broader semantic
  understanding, previous methods \cite{VDI, zheng2014vtg} applied large-scale
  vision-language models like CLIP \cite{clip}, InternVideo
  \cite{wang2022internvideo} or BLIP2 \cite{blip2}, but they still lack the ability to localise fine-grained
  moment-text associations due to the pre-training on coarse
  (holistic broad-strokes weakly-supervised) image-text
  or video-text pairs. Although unsupervised methods~\cite{psvl,dscnet,spl} were
  proposed to generate pseudo moment-text associations from unlabelled
  videos and use them to train fully supervised VMRs, they are
    inherently limited by both the quantities of videos available and
    error-propagation from self-labelling. In light of these challenges,
  contemporary cross-domain solutions are either impractical due to
  insufficient training videos or suboptimal in capturing fine-grained
  associations. Our method is more scalable and optimised for learning novel
    target semantic associations from only target text query without
    any videos from the target domain.

\noindent \textbf{Video Diffusion Models.}
Current video diffusion models have demonstrated promising outcomes in
the domains of video synthesising \cite{cogvideo,TAV,feng2023ccedit}. We primarily focus on \sg{two
specific perspectives}: \dz{video generation and video editing. }

\noindent \textbf{Video Generation.} 
\dz{ Due to \sg{the difficulties
  in collecting high-quality video data}, existing video generation
  methods \sg{leverage both} images and videos
  \sg{in model training}
  \cite{makeavideo,videoldm,wang2023modelscope}. \sg{Moreover},
  predominantly trained on images (2.3B images v.s. 10M videos),  they
  are \sg{incapable of generating fine details in} human-centred videos of complex dynamics. 
\sg{Further}, they rely solely on text prompts and \sg{are `blind' to
  visual controls from each subject instance of an action and the scene
  context~(background environment)}.

To generate videos with better dynamics, motion customisation methods
\cite{zhao2023motiondirector,dreamvideo} \sg{concentrated on learning} a specific motion with given samples, which requires additional labelled text-video pairs for each action. To generate videos with \sg{specified subjects},
Videobooth \cite{jiang2024videobooth} proposed to input image as a prompt, however, it is still insufficient to handle human instances with rich details (shown in the Supplementary). 
 Dreamix \cite{molad2023dreamix}  introduced a subject-driven action generation by a mixed reconstructing strategy of subject-driven image generation \cite{ruiz2023dreambooth,customization_dreambooth,chen2023disenbooth} and video generation \cite{TAV}. 
However, the mixed training of reconstructing images and actions \cite{molad2023dreamix} is likely to entangle the two features in latent embedding, resulting in inaccurate specific information or overfitting on the source action \cite{chen2023disenbooth}.
 This highlights the need to design a better subject-preserving video editing method. 
}

\dz{\noindent \textbf{Video Editing.}}
Existing methods \cite{TAV,groundavideo,yan2023magicprop,liew2023magicedit} have demonstrated proficient object editing capabilities by manipulating the associated textual descriptions. To 
maintain the integrity of the background and 
prevent alterations to regions not in the focus on change, plug-and-play techniques \cite{qi2023fatezero,videop2p} employed a decoder with cross-attention derived masks to protect unrelated areas, necessitating users to interactively discern and selectively specify the interchangeable parts between the source and target prompts. 
Other methods \cite{geyer2023tokenflow,lu2023fuse,yang2023rerender} were proposed to eliminate the training process on a specific video and to edit video objects directly using priors from image diffusion models \cite{stable-diffusion}.
\dz{However, current video editing} methods fall short in the realm
  of VMR video simulation, particularly in crafting moments with a
  consistent subject engaging in various actions. These methods lack
  the necessary controls to edit actions while preserving the
  distinctive features of the subject. VMR tasks typically
  require distinguishing subtle differences between matching and
  non-matching video moments, and they often feature the same subjects,
  backgrounds, and video styles. Therefore, the aforementioned
  shortcomings of existing \dz{video editing methods} restrict
    inherently generating meaningfully diverse and visually plausible
    data for learning novel unseen concepts in VMR.

  Overall, existing generative models \cite{videop2p,qi2023fatezero} heavily rely on designing a delicate generation control,  
\eg through an interactive selection of a target prompt, to produce plausible videos through trial-and-error. How to design effective automatic
controls for generating target moment-text associations capable 
of optimising novel semantic cross-domain video moment retrieval model 
learning has not been studied, nor it is straightforward.

\section{Method}

Our aim is to simulate video moment retrieval (VMR) training data with
fine-grained moment-text associations, and autonomously regulate
  the video generation process using a collection of sentences,
  without any human intervention by interactive text prompts or
  reliance on target exemplar videos. In this section, we first recap diffusion models, then present our Fine-grained Video Editing framework (\abbrname) with an
instance-preserving action editing model (Fig. \ref{fig:method}) and an automatic video generation and hybrid selection
pipeline (Fig. \ref{fig:method_2}). 
\subsection{Latent Diffusion and DDIM Inversion}
\noindent \textbf{Latent Diffusion Models (LDMs).}
LDMs are introduced to diffuse and denoise data \cite{diffuionmodel} within a compressed latent space. For an image $x$,
the process starts with the encoding of  $x$ into a latent representation \cite{vae}: $z = \mathcal{E}(x)$.
Gaussian noise is then added to this representation to create \( z_t \) at timestep \( t \). A denoising autoencoder is subsequently trained to predict the Gaussian noise in the latent representation, aiming to reverse the noise addition. The objective is defined as:

\begin{equation}
L_{LDM} = \mathbb{E}_{\mathcal{E}(x), \epsilon \sim \mathcal{N}(0, I), t}\Big[\| \epsilon - \epsilon_\theta(z_{t}, t, p) \|_{2}^{2}\Big],
\end{equation}
where \( \epsilon_\theta \) is a U-Net \cite{unet} architecture 
conditioned on a timestep $t$ and a text prompt embedding $p$ and the visual input $z_t$.

\noindent \textbf{DDIM Inversion.} 
DDIM Inversion \cite{ddim} maps a clean latent representation \( z \) to its noisy counterpart using a sequence of reverse timesteps from $t = T-1$ to $1$. The  DDIM iterative process is defined by:

\begin{equation}
    \hat{z}_t = \sqrt{\alpha_t} \hat{z}_{t-1} + \left( \sqrt{1 - \alpha_t} - \sqrt{\frac{1 - \alpha_t}{\alpha_{t-1}}} \right) \epsilon_{\theta},
\end{equation}
where \( \hat{z}_t \) denotes the estimated noisy latent state at timestep \( t \) and \( \alpha_t \) represents the variance schedule at timestep \( t \). 
 
\subsection{Instance-Preserving Action Editing}

To simulate a VMR video involving the same subject executing different actions, we aim to address the limitations of current \dz{video} editing methods \cite{TAV, videop2p}, which lack controls for constraining subject specifics. Additionally, we aim to overcome the inaccuracies in instance information learned by subject-driven video generation methods \cite{molad2023dreamix}, attributed to their mixed subject-specific and motion training strategy. Specifically, we separate the learning of subject-specific instance information and the video motion (Fig. \ref{fig:method}). We first train an image model to align a text token `[v] person' with the visual information of the shared subject instance across frames. This alignment is achieved in the 2D-attention layer containing self-attention (visual) and cross-attention (textual-visual) layers. Subsequently, we \textit{freeze} the learned textual-visual alignment and introduce a temporal layer following the 2D-attention layer to capture the video motion.
\begin{figure}[t]
    \centering
    \includegraphics[width=1\linewidth]{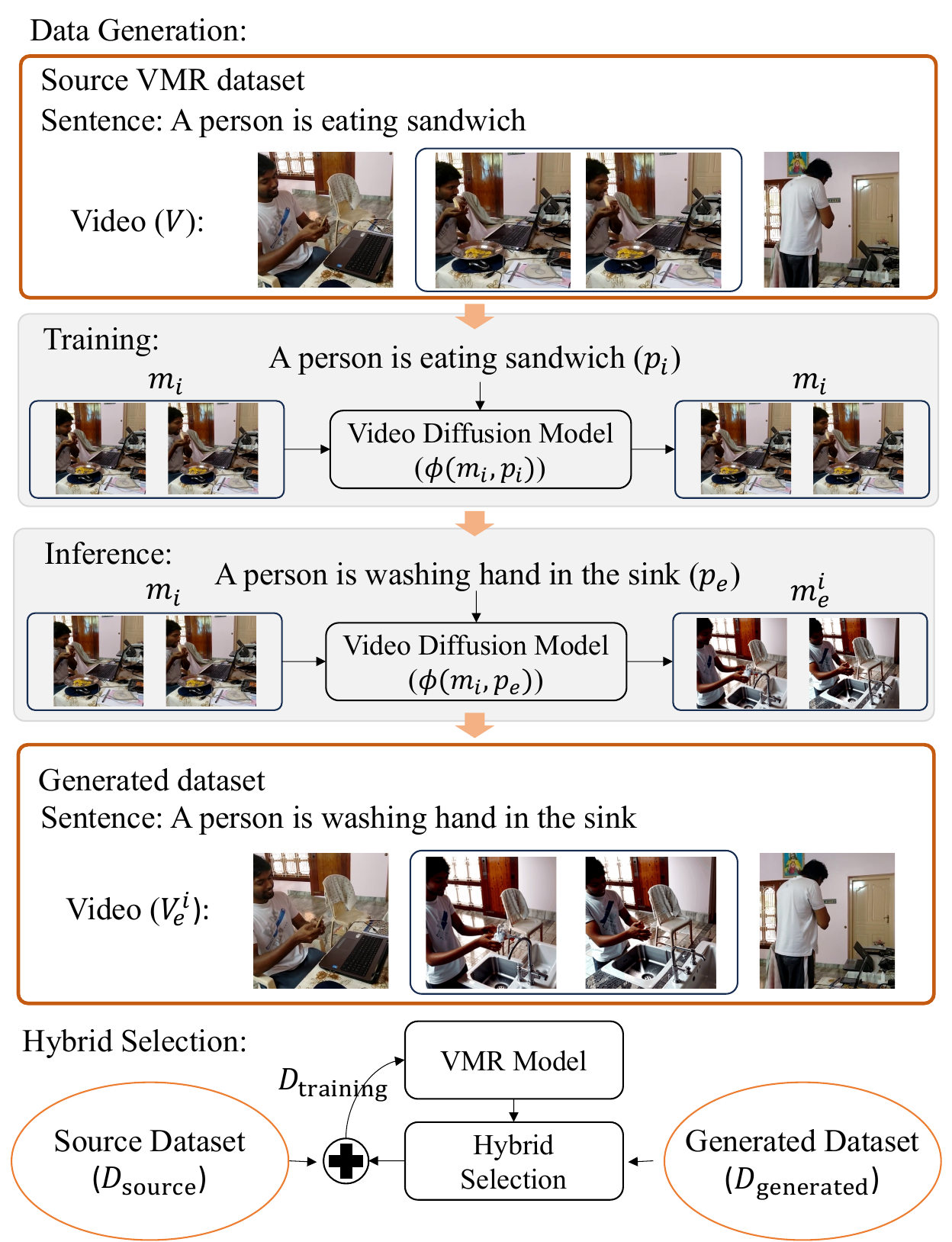}
    \caption{Data generation and hybrid selection. For data generation, we first train the video diffusion model $\phi$ to align moment $m_i$ with a sentence $p_i$, then we use an editing prompt $p_e$ to edit the moment to $m_e^i$. The hybrid selection strategy includes a cross-modal relevance and uni-modal structure score to select high-quality generation, as well as a model performance disparity to select beneficial data for VMR training.}
    \label{fig:method_2}
    \vspace{-10pt}

\end{figure}

\noindent \textbf{Stage1: Instance Descriptor Learning.}
In the first stage, we take the video as a set of unordered frames for instance descriptor learning. For effective learning, we select a subset of frames from the whole frame set, aiming to maximize diversity whilst ensuring frame clarity by minimising noise such as motion blur.
Mathematically, given a frame \( f_i \) and its immediate neighbouring frames \( f_{i-1} \) and \( f_{i+1} \), we seek frames to maximize the following function:
\begin{equation}
\Phi(f_i) = \delta(f_i, f_{i-1}) + \delta(f_i, f_{i+1}) + \chi(f_i),
\end{equation}
where \( \delta \) denotes the dissimilarity measure between frames using histogram, and \( \chi \) signifies the frame's clarity, determined with the Laplacian operator to evaluate the visual sharpness. We select 10 frames with higher  $\Phi(f_i)$ scores.

For the training of the instance descriptor, we leverage the Dreambooth strategy \cite{ruiz2023dreambooth} to train an image diffusion model to reconstruct the selected video frames from a text token `[v] person', while simultaneously reconstructing images of other instances based on the prompt `a person'. The model's goal is to embed the visual instance within the output domain specified by the text token `[v] person'. The objective \(L_{IDL}\) is formulated as:
\begin{equation}
\begin{split}
& L_{IDL}  =  \mathbb{E}_{\mathcal{E}(x_{inst}), \epsilon \sim \mathcal{N}(0, I), t}\Big[\| \epsilon - \epsilon_\theta(z_{t,inst}, t, p_{inst}) \|_{2}^{2}\Big], \\
& + \mathbb{E}_{\mathcal{E}(x_{class}), \epsilon \sim \mathcal{N}(0, I), t}\Big[\| \epsilon - \epsilon_\theta(z_{t,class}, t, p_{class}) \|_{2}^{2}\Big],
\end{split}
\end{equation}
where \(x_{inst}\) and \(x_{class}\) denote the input of instance image and class images (other instances in the same class), \(p_{inst}\) and \(p_{class}\) denote their corresponding prompts. 
After training, the text token is aligned with instance visual information, enabling a generalisable generation by combining it with other concepts.

\noindent \textbf{Stage 2: Temporal Encoding.} When editing a moment based on a target sentence, it is crucial to preserve the original unrelated motions not referenced by the sentence. This includes maintaining subject-specifics, background, and video style intact. As shown in Fig. \ref{fig:method}, we append a temporal attention layer after each 2D-attention layer. In order to fix the alignment of `[v] person' with the visual content, in contrast to the methods employed in Dreamix \cite{molad2023dreamix}, we freeze previously optimised layers and only train the temporal layer. 
The training loss $L_{TE}$ is formulated as:

\begin{equation}
\begin{split}
 & L_{TE}(m,p) = ||m-\phi(m,p) || \\ 
 & = \frac{1}{n} \sum_{i=1}^{n}  \mathbb{E}_{\mathcal{E}(f_i), \epsilon \sim \mathcal{N}(0, I), t} \Big[\| \epsilon - \epsilon_\gamma(z_{t,i}, t, p) \|_{2}^{2}\Big],
\end{split}
\end{equation}
where $n$ is the frame number in moment $m$, and each frame is denoted as $f_i$. $\phi$ is our model consisting of a U-Net architecture ($\epsilon_\gamma$) with a temporal layer attached after each 2D-attention layer. $p$ is the prompt describing the action.

\subsection{Data Generation and Hybrid Selection}
\label{sec:3.3}

\subsubsection{Data Generation.}
Considering a video $V$, a VMR dataset \cite{charades} will provide a list of moments in the video with their sentence descriptions,  denoted as $\{(m_i, p_i)\}_{i=1}^{a}$, where $m_i$ denotes the $i^{th}$ moment, $p_i$ denotes the corresponding description and $a$ is the number of moments. An editing sentence is also provided, represented as $p_e$. As depicted in Fig.~\ref{fig:method_2}, our model involves a training and inference stage: 
we first train our video diffusion model on each moment-text pair as $L_{TE}(m_i, p_i)$, aligning the moment $m_i$ with its textual description $p_i$. Then we modify the specific moment $m_i$ in the video $V$ using the sentence \(p_e\), as:
\begin{equation}
m^i_e = \phi(m_i, p_e),
\end{equation}
where $m^i_e$ denotes the $p_e$-edited version of moment $m_i$. Then $m^i_e$ replaces the original moment $m_i$ to create a new variant of the video, denoted as $V^i_e$. For the video $V$ with $a$ moments, we generate $a$ different video variants \(\{V^1_e, V^2_e, \ldots, V^a_e\}\), where \(V^i_e\) contains all original moments except for the $i^{th}$ moment, which is replaced by its edited version \(m^i_e\), resulting in a set of videos where each variant showcases a unique modification at a 
distinct moment.

\subsubsection{Hybrid Selection.}
Without a delicate selection of the editing prompt, automatic VMR data generation may result in noisy or repetitive videos.
 In order to select high-quality and beneficial data for VMR training, we design a hybrid selection strategy with three quantitative metrics: cross-modal relevance, uni-modal structure, and model performance disparity. 
 For cross-modal relevance,
   we notice that a lack of semantic relevance between
   the source video and the target sentence might result in
   implausible outcomes where the generated video content does
   not align well with the provided text. Training a VMR model using
 noisy pseudo moment-query pairs misleads the model to
   learn inaccurate moment-text associations. To this end, we introduce a cross-modal relevance score to evaluate the
   coherence between the target prompt and the generated video
   moment:
\begin{equation}
s_{c}(p_e,m_e)=\frac{1}{N} \sum_{i=1}^{N} \cos(\text{VLM}(p_e),\text{VLM}(f_{m_e}^i)),
\end{equation}
where $s_c$ denotes the cross-modal relevance score, $p_e$ the editing prompt, $f_{m_e}^i$ the $i^{th}$ frame in the generated (edited) moment $m_e$, and $N$ the frame number in $m_e$. VLM denotes a vision-language model pre-trained on large-scale datasets.

 For uni-modal structure score, we evaluate the visual consistency between the source and generated video moment:
\begin{equation}
s_{u}(m_s,m_e)=\frac{1}{N} \sum_{i=1}^{N} \cos(\text{VM}(f_{m_s}^i),\text{VM}(f_{m_e}^i)),
\end{equation}
where $s_u$ is the uni-modal structure score, $f_{m_s}^i$ is the
$i^{th}$ frame in the source moment $m_s$. VM is a visual encoder
pre-trained on a large-scale vision dataset, which can predict the general structure, including the environment and object features, of an image. Successful and high-quality editing is the generation that both matches the text prompt and maintains the original video structure,
so we integrate the two metrics using a harmonic score: 
\begin{equation}
s_{cu}(p_e,m_e,m_s) =\frac{ 2 \times s_{c}(p_e,m_e) \times s_{u}(m_s,m_e)} {s_{c}(p_e,m_e) + s_{u}(m_s,m_e)}.
\end{equation}
 The selection process is represented as follows:
\begin{equation}
D_{cu} = \text{TOP}_k(\{ (d, s_{cu}(p_e,d,m_s)) \mid d \in D_\text{generated} \}),
\end{equation}
where $D_\text{generated}$ is the generated dataset, $D_{cu}$ comprises $k$ samples with top $s_{cu}$.

In addition to the high-quality moment selection, we underscore the significance of enriching VMR
  training by diverse data that is both visually plausible and
    semantically meaningful for enhancing VMR novel concept
    generalisation, rather than simply duplicating existing content
  from the source domain. To address this, we introduce the
  third metric, called model performance disparity. This metric
  measures the degree to which the model's predictions differ from the
  ground truth or desired outcomes across various samples. Higher
  model error rates in certain samples indicate instances where the
  model struggles to accurately capture the relationship between
  moments and queries. These samples are earmarked for further
  analysis or refinement in the training process. In practice, we
  evaluate VMR on the previously filtered dataset $D_{cu}$ and
  incorporate only those samples of high disparity in training:

\begin{equation}
\begin{split}
& D_\text{mpd} = \text{TOP}_l (\{ (d, -\text{VMR}(d)) \mid d \in D_{cu} \}), \\
& D_{\text{training}} = D_{\text{source}} \cup D_\text{mpd},
\end{split}
\end{equation}
where $D_{\text{training}}$ comprises the source data $D_{\text{source}}$  and additional $D_\text{mpd}$ data with a length of $l$, selected  with low VMR performance.
This enables us to identify and select cases that are not adequately handled by the existing model.

  \begin{table}
    \setlength{\tabcolsep}{1.4mm}

  \centering

  \begin{tabular}{l|c|c|ccc}
    \hline
     \multirow{2}{*}{Method}&\multirow{2}{*}{Year} &\multirow{2}{*}{Target}&\multicolumn{3}{c}{Charades-STA} \\
         \cline{4-6}
     &&&R1@0.5&R1@0.7 & mIoU  \\
    \hline
    EVA  \tablefootnote{ \label{f1}These methods are not reported on the same split.} &2022&\multirow{2}{*}{\shortstack{Video\\\& Text}}&
    40.21 &18.77&-\\
    MMCDA \textsuperscript{\ref {f1}}&2022&& 54.80&35.77&-\\
    \hline
\multicolumn{6}{c}{I3D feature}     \\
    \hline
        TMN  &2018&\multirow{7}{*}{No} &9.43 & 4.96 & 11.23  \\
    TSP-PRL&2020&&14.83 & 2.61 &14.03 \\

    2D-TAN   &2020& &29.36 &13.12 &28.47   \\
    LGI  &2020&  &  26.48 & 12.47 & 27.62 \\
      VSLNet  &2020 & & 25.60 &10.07 & 30.21\\
    VISA &2022&  &42.35 & 20.88 & 40.18 \\

                               VDI$^\dagger$ &2023  &&46.19 &26.19 & 40.95 \\
                              \hline \abbrname (Ours)&2025&Text&\textbf{48.51}&\textbf{28.48}&\textbf{42.67} \\

        \hline
        \multicolumn{6}{c}{Slowfast feature}     \\
    \hline
    M-DETR &2021&\multirow{4}{*}{No} &43.45&21.73&38.37 \\
    QD-DETR &2023&&48.20&26.19&43.22 \\
     UVCOM$^\dagger$  &2024& &48.63&28.57&42.65\\
     MESM$^\dagger$ &2024   &  & 51.08&29.78&44.16\\
           \hline
           \abbrname (Ours) &2025&Text & \textbf{52.37} &\textbf{31.94}&\textbf{44.59}\\
           \hline

  \end{tabular}
    \caption{Novel-word testing on Charades-STA. The `Target'
      column indicates the information required from the
      target domain. \dz{Symbol `$\dagger$' indicates our implementation with the author-released code. }}
  \label{table:novel word charades}

\end{table}

\section{Experiments}
To evaluate the effectiveness of our Fine-grained Video Editing framework (\abbrname), we validate on both video moment
retrieval and video action editing tasks.

\subsection{Video Moment Retrieval}

\noindent \textbf{Data Setup.} 
\dz{To assess \abbrname for novel
semantic VMR, we employed the `novel-word' split
\cite{visa} on Charades-STA \cite{charades}, where the
testing split contains `novel-words' not seen in the training
split.
For QVHighlights \cite{QVHIGHTLIGHTS} and TaCoS \cite{Tacos}, we sample sentences from the standard training split and exclude them from the training set.
In our
implementation, we selected 50/300/300 sentences separately from each
dataset for data generation.
Selection details are shown in \sg{the} Supplementary.

\noindent \textbf{Implementational Details.}
For each target sentence, we created 100/50/50 videos for each dataset. This resulted in a total of 5,000/15,000/15,000 generated videos, ready to be chosen to support the training of the VMR model. For hybrid selection, we used CLIP~\cite{clip} to compute the cross-modal relevance score and DINO~\cite{DINO} for the uni-modal structure score. We set $k$ to 500, 1500 and 1500 respectively for the three datasets. For the model performance disparity metric, we set $l$ to be 100, 500  and 500 respectively for each dataset. We adopted R1@µ, mAP@µ, mIoU, and mAPavg as the evaluation metrics.}

\begin{table}[h!]
      \setlength{\tabcolsep}{1.1mm}
  \centering
  \begin{tabular}{l|c|cc|ccc}
    \hline
    \multirow{3}{*}{Method} &    \multirow{3}{*}{Target} 

 &           
\multicolumn{5}{c}{QVHighlights} \\
         \cline{3-7}

&&\multicolumn{2}{c|}{R1}&\multicolumn{3}{c}{mAP} \\
&&@0.5&@0.7&@0.5&@0.75&avg \\
    \hline
  
M-DETR &\multirow{8}{*}{\shortstack{Video\\\& Text}}&52.89&33.02&54.82&29.40&30.73 \\
UMT & &56.23&41.18&53.38&37.01&36.12 \\
UniVTG& &58.86&40.86&57.60&35.59&35.47 \\
MH-DETR& & 60.05&42.28&60.75&38.13&38.38 \\
QD-DETR& & 63.06&45.10&63.04&40.10&40.19 \\
EaTR & &61.36&45.79&61.86&41.91&41.74 \\
MESM & &62.78&45.20&62.64&41.45&40.68 \\
UVCOM & & 63.55&47.47&63.37&42.67&43.18 \\
\hline
MESM$^\dagger$ & \multirow{2}{*}{No} & 61.95&45.03&60.23&38.94&39.03 \\
UVCOM$^\dagger$ & & 61.39&45.45&60.43&40.38 &40.30\\
\hline
FVE (Ours) & Text&\textbf{63.35}&\textbf{47.16}&\textbf{62.17}&\textbf{42.00}&\textbf{41.33} \\
\hline
  \end{tabular}
    \caption{\dz{VMR results on QVHighlights. The `Target' column and the symbol `$\dagger$' denotes the same as Table \ref{table:novel word charades}.}}

  \label{table:qvhilight}
\end{table}
\begin{table}[h!]

  \setlength{\tabcolsep}{0.2cm}
  \centering
  \begin{tabular}{l|c|ccc}
    \hline
    \multirow{2}{*}{Method}& \multirow{2}{*}{Target}  &\multicolumn{3}{c}{TaCoS} \\
             \cline{3-5}

     &&R1@0.3&R1@0.5&mIoU \\
    \hline
    VSLNet &\multirow{9}{*}{\shortstack{Video\\\& Text}}&29.61&24.27&24.11 \\
    2D-TAN &&37.29&25.32&- \\
    CBLN &&38.98&27.65&- \\
    RaNet &&43.34&33.54&- \\
    SeqPAN&&31.72&27.19&25.86 \\
    SMIN&&48.01&35.24&-\\
    MMN & &39.24&26.17 &- \\
    MS-DETR& &47.66&37.36&35.09 \\
    MESM & &52.69&39.52&36.94 \\
    \hline
        MESM$^\dagger$ &No& 44.01&29.39&29.15 \\
\hline
    FVE (Ours)&Text&\textbf{48.09}&\textbf{31.92}&\textbf{31.61} \\
    \hline

  \end{tabular}
    \caption{\dz{VMR results on TaCoS. The `Target' column and the symbol `$\dagger$' denotes the same as Table \ref{table:novel word charades}}.}
  \label{table:tacos}
\end{table}

\dz{\subsubsection{Comparisons.}

We compare our \abbrname with the following methods: EVA \cite{eva}, MMCDA \cite{mmcda}, TMN \cite{tmn2018}, TSP-PRL \cite{tsp}, 2D-TAN \cite{2dtan}, LGI \cite{lgi2020}, VSLNet \cite{vslnet}, VISA \cite{visa}, VDI \cite{VDI}, M-DETR \cite{QVHIGHTLIGHTS}, QD-DETR \cite{qd-detr}, UVCOM \cite{UVCOM}, MESM \cite{MESM}, UMT \cite{umt}, UniVTG \cite{lin2023univtg}, MH-DETR \cite{mhdetr}, EaTR \cite{EaTR}, CBLN \cite{cbln}, RaNet \cite{ranet}, SeqPAN \cite{seqpan}, SMIN \cite{SMIN} and MS-DETR \cite{ms-detr}.

For Charades-STA, 
we apply our method on two different feature extractors, I3D \cite{i3d} and Slowfast \cite{slowfast} using VDI \cite{VDI} and MESM \cite{MESM} as the baseline separately. 
As shown in Table \ref{table:novel word charades}, with a
 collection of 50 out of 703 sentences, we improve the performance for
 Charades-STA from 29.78\% to 31.94\% on R1@0.7 and we reach the SOTA on all metrics.
 For QVHighlights,
 we take Slowfast as the feature extractor and UVCOM as the baseline.
As shown in Table \ref{table:qvhilight}, we obtain gains in all metrics over the baseline model UVCOM. Also, with only a collection of text, we reach comparable performance to those requiring video\&text pairs (row 8 vs. row 11 on R1@0.5 and R1@0.7). For TaCoS, we apply C3D~\cite{c3d} as the feature extractor and MESM as the baseline.
As shown in Table \ref{table:tacos}, we also obtain gains in all metrics on over the baseline model MESM.
}\dz{More comparisons 
     are \sg{given} in the \sg{Supplementary}.} 
\begin{table}[h!]

  \setlength{\tabcolsep}{0.2cm}
  \centering

  \begin{tabular}{l|c|ccc}
    \hline
    \multirow{1}{*}{Method} 
 &           
\multirow{1}{*}{Datasize} &\multirow{1}{*}{R1@0.5}&\multirow{1}{*}{R1@0.7} &mIoU \\
    \hline
    No & 3533&46.19&26.19&40.95\\
    \hline
        Concat & 4033&46.23&26.51&40.17\\

    \hline
    
    $R_1$& \multirow{3}{*}{4033}&47.19&27.19&41.05\\
    $R_2$& &45.61&26.33&40.40\\
    $R_3$& &44.03&25.47&40.00\\
    \hline 
    \abbrname (Ours) & 4033&\textbf{48.51}&\textbf{28.48}&\textbf{42.67}\\
    \hline

  \end{tabular}
    \caption{Ablation on the effect of data volume. `No' denotes no use of generated data, `Concat' video generation through the random concatenation of existing videos.  `$R_1$'-`$R_3$' a random sampling with different random seeds,   `Datasize' denotes the number of videos for each method.}
  \label{table:novel_word_discuss_size}
\end{table}

  \subsubsection{Ablation Study.}
  
\dz{
 In this study, we first 
   eliminate the effect of data volume and then highlight the importance of data selection. We conduct ablation studies using I3D features on the Charades-STA dataset and model VDI. } As shown in
Table~\ref{table:novel_word_discuss_size},  increasing the data volume
by adding replicated samples does not enhance performance~(row
1 vs. row 2). Moreover, in comparison to randomly sampling data from
the generated pool (`$R_1$'-`$R_3$') into the training set, with an equal volume of data, \abbrname
selects effectively those generated videos that enhance VMR
  model training.

Table \ref{table:hybrid-ablation} shows an ablation study on
  the two hyperparameters for the number selected from: the harmonic score between cross-modal
  relevance and the uni-modal structure score ($k$) and the model
  performance disparity ($l$). We observe the best combination is
  $k$=500 and $l$=100. 
  \dz{More ablation studies including the combination of $s_c$ and $s_u$ scores, the ablation of frame selection, the ablation of the location to replace the frames, and ablations on other datasets are presented in the Supplementary.}

\begin{table}[t]
  \centering
  \setlength{\tabcolsep}{0.15cm}

  \begin{tabular}{ll|ll|ccc }
    \hline
 $s_{cu}$&$k$&$s_{mpd}$  &$l$ &R1@0.5 &R1@ 0.7&mIoU \\
    \hline
    $\times$&1500&$\times$&- &43.88&25.04&38.50\\
    \hline
      \multirow{4}{*}{\checkmark} & 1500 &\multirow{4}{*}{$\times$}&\multirow{4}{*}{-} 
      &44.89&26.19&40.79\\
       &1000& &&43.60&24.46&40.05\\
     &500& & &46.64&27.16&41.21 \\
       &100&&&46.91&\textbf{28.78}&41.14\\
      \hline
      \multirow{3}{*}{\checkmark}&\multirow{3}{*}{500}   &\multirow{3}{*}{\checkmark} & 200 & 45.75&26.04&40.88\\

       &&&100&\textbf{48.51}&28.48&\textbf{42.67}\\
       &&&50& 46.19&25.47&41.35\\
       \hline
 \multirow{2}{*}{\checkmark}&\multirow{2}{*}{100}   &\multirow{2}{*}{\checkmark} & 50&47.34&28.49&42.24 \\
        &&&10 &47.48&27.05&41.89\\

 \hline 
  \end{tabular}
    \caption{ Ablation of the hybrid selection. $s_{cu}$ and $k$ denote the score of combining cross-modal relevance and uni-modal structure and its selecting numbers. $s_{mpd}$ denotes model performance disparity selection with a number $l$. $s_{cu} = \times$ denotes a random sampling from the generated data. $s_{mpd} =\times $ denotes $s_{mpd}$  is not applied and we use all the samples selected by $s_{cu}$. }
  
  \label{table:hybrid-ablation}

\end{table}

\subsection{Action Editing}
For action editing, we compare with Tune-A-Video \cite{TAV}, Video-P2P \cite{videop2p}, Fatezero \cite{qi2023fatezero} and Dreamix \cite{molad2023dreamix}.
We collect a 10-video dataset with videos from Charades-STA and videos in the
wild. We carry out comparisons on both quantitative and qualitative evaluations.

\begin{figure}
    \centering
    \includegraphics[width=1\linewidth]{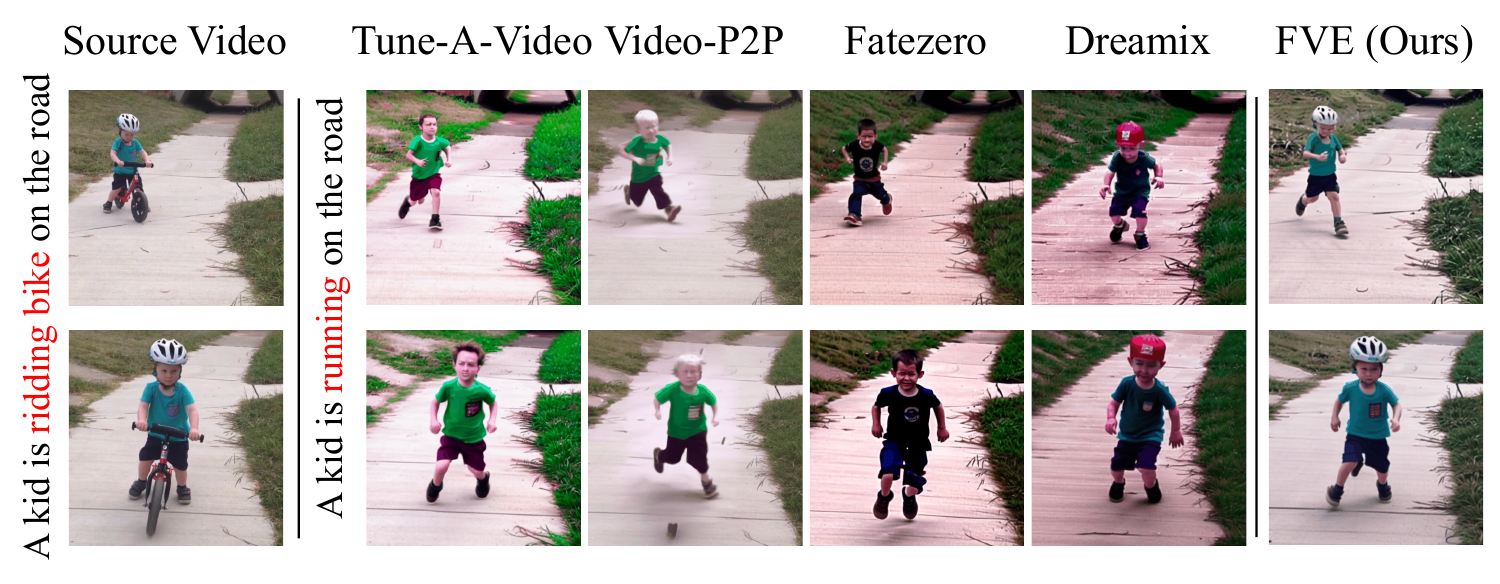}
    \caption{Qualitative comparisons.
    The first and last frames of the video are presented. 
    }
    \label{fig:qualitative}
    \vspace{-10pt}

\end{figure}

For quantitative evaluation, we evaluate the result using cross-modal relevance ($s_c$) and uni-modal structure scores ($s_u$). We define a successful generation as a video that optimizes both metrics quantified by their harmonic score $s_{cu}$. As shown in Table \ref{table:quan}, previous methods either fail to maintain the video structure (TAV, Video-P2P) or show inferiority in cross-modal relevance (Dreamix, Fatezero), \abbrname demonstrates the best result with the harmonic score that combines
the impact of both metrics.
\dz{We present the qualitative evaluation of a single video due to space limitations. }
As depicted in Fig. \ref{fig:qualitative}, 
\abbrname generates videos that adeptly preserve instance appearance and generalise to new actions. \dz{More visualisations are presented in the Supplementary.}

\begin{table}[t]
  \centering
 
  \setlength{\tabcolsep}{0.4cm}
  \begin{tabular}{l|c|c|c}
    \hline
    Method & $s_c$ &$s_u$&$s_{cu}$   \\
    \hline
    Tuen-A-Video  & \textbf{0.2910}&0.4939&0.3641\\
    Video-P2P &0.2895&0.5802&0.3862 \\
 
    Fatezero &0.2621&0.7061&0.3822\\
    Dreamix &0.2672&0.6826&0.3841\\

 \hline 
 \abbrname (Ours)&0.2722&\textbf{0.7086}&\textbf{0.3933} \\
 \hline 
  \end{tabular}
   \caption{ Quantitative comparisons. $s_c$ denotes the cross-modal relevance score, $s_u$ the  uni-modal structure score and $s_{cu}$ their harmonic score.}
  
  \label{table:quan}
\end{table}

\section{Conclusion}

  In this work, we addressed the problem of unseen novel semantic
  video moment retrieval (VMR) cross domains without having seen any target videos in
  training. Given the aim of learning
a target domain by only text sentences describing new concepts, we
proposed a Fine-grained Video Editing framework (\abbrname) to edit
source videos automatically controlled by target sentences to
simulate target domain training data. To control the
generation process and select both visually plausible and
  semantically meaningful fine-grained video hypotheses for VMR training, we
formulated an instance-preserving video diffusion model and
a hybrid data
selection strategy. \dz{ Experimental results on three datasets demonstrated the effectiveness and generality of our method improving performance on VMR. Evaluation of video editing further demonstrated the ability of our method to change the action in a video and maintain the subject information.}
Future directions could explore long-temporal video editing conditioned on complex sentence prompts.

\section*{Acknowledgements}

This work was partially supported by Adobe, Veritone, NSFC (62372014), CSC, and Queen Mary University of London’s Apocrita HPC facility from QMUL RESEARCH-IT.
{
\bibliography{aaai25}
}

\end{document}